\documentclass[letterpaper,journal]{IEEEtran}
\usepackage{amsmath,amsfonts}
\usepackage{algorithmic}
\usepackage{algorithm}
\usepackage{array}
\usepackage[caption=false,font=normalsize,labelfont=sf,textfont=sf]{subfig}
\usepackage{textcomp}
\usepackage{stfloats}
\usepackage{url}
\usepackage{verbatim}
\usepackage{graphicx}
\usepackage{cite}
\usepackage{hyperref}
\usepackage{booktabs} % 更好的三线表
\usepackage{multirow} % 合并表格单元格
\usepackage{xcolor}

\begin{document}
\bstctlcite{IEEEexample:BSTcontrol}

\title{VLBM: Variational Latent Basis Modeling for OOD Robust Multivariate Time Series Forecasting}

\author{Xudong~Zhang\textsuperscript{1,2,3},
Jierui~Lei\textsuperscript{4},
Jiacheng~Li\textsuperscript{5},
Lingdong~Shen\textsuperscript{5},
Jian~Cui\textsuperscript{3},
and~Haina~Tang\textsuperscript{1}%
\thanks{Xudong Zhang and Jierui Lei contributed equally to this work. Haina Tang is the corresponding author.}%
\thanks{\textsuperscript{1}School of Artificial Intelligence, University of Chinese Academy of Sciences, Beijing 100049, China.}%
\thanks{\textsuperscript{2}Center for Machine Learning Research, Peking University, Beijing, China.}%
\thanks{\textsuperscript{3}Amap, Alibaba Group, China.}%
\thanks{\textsuperscript{4}School of Advanced Interdisciplinary Sciences, University of Chinese Academy of Sciences, Beijing, China.}%
\thanks{\textsuperscript{5}Environmental Microbiome and Innovative Genomics Laboratory, Peking University, Beijing, China.}%
\thanks{Haina Tang is with the School of Artificial Intelligence, University of Chinese Academy of Sciences, Beijing 100049, China (e-mail: hntang@ucas.ac.cn).}%
% \thanks{Manuscript received Month Day, Year; revised Month Day, Year.}
}



\maketitle

\begin{abstract}
Out of distribution (OOD) events in multivariate time series forecasting are rare but often dominate real world risk, making average case forecasting insufficient for reliable deployment. Under standard average risk training on mixed ID/OOD distributions, optimization signals from rare OOD events can be overwhelmed by frequent in distribution (ID) patterns, so strong benchmark accuracy may not translate into reliability under high impact shifts. To address this issue, we propose VLBM (Variational Latent Basis Model), a theory guided latent forecasting framework that separates stable dynamics from OOD induced deviations. VLBM learns a shared latent basis that defines a low rank subspace for stable ID dynamics, explicitly decomposes inputs into basis subspace components and orthogonal residual components, and aligns a future aware posterior with a future blind prior so that test time latent inference depends only on historical input. Across 12 benchmark tasks spanning transportation, weather, power systems, and other real world domains, including newly constructed real world OOD traffic datasets, VLBM achieves state of the art OOD robustness and ID accuracy, with average MAE and MSE gains of 15.08\% and 7.74\% over the strongest baseline. On a synthetic simulation dataset, VLBM also consistently achieves the best performance and better tracks OOD pulse recovery. These results support latent structured forecasting as a principled route to robust prediction under mixed ID and OOD conditions. The code is available at \href{https://github.com/leijieruilq/VLBM_OOD_forecast}{https://github.com/leijieruilq/VLBM\_OOD\_forecast}.
\end{abstract}


\section{Introduction}
\begin{figure*}[t]
    \centering
    \includegraphics[width=0.9\textwidth]{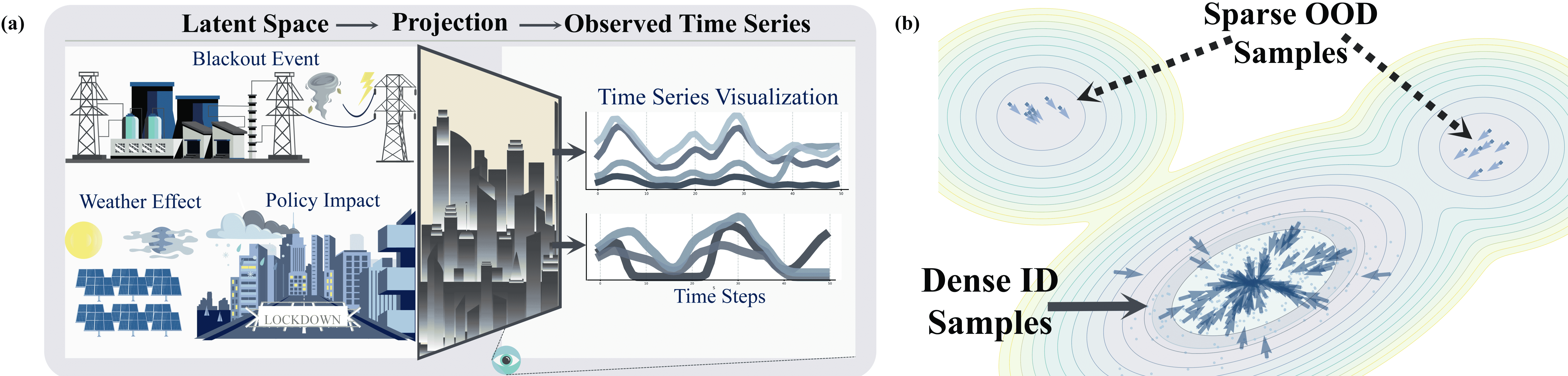}
    \caption{
    Conceptual overview of OOD robust forecasting under mixed ID/OOD conditions.
    (a) Observed time series are partial projections of latent system dynamics.
    (b) Under standard average risk training on mixed ID/OOD distributions, frequent ID patterns dominate optimization and representation learning, while rare OOD signals have limited influence.
    }
    \label{fig:overview}
\end{figure*}
Despite the widespread success of multivariate time series forecasting in
transportation, energy, and environmental systems
\cite{jin2024survey,zhang2024self,Timedart,lei2025transformer},
most existing models operate in observation space,
learning future predictions directly from historical signals,
with limited access to the underlying dynamics that generate them.
In the spirit of Plato's Allegory of the Cave \cite{PlatoRepublic}, the
observed sequence can be viewed as a projection of hidden causes, leaving the
full dynamical process only partially observed. This view motivates a concrete forecasting
question: how can a model recover stable latent dynamics from surface level
observations while remaining responsive to rare deviations
(Fig.~\ref{fig:overview}(a))?

This mismatch becomes particularly harmful under OOD conditions.
Events such as traffic accidents, holidays, extreme weather, and human
interventions \cite{Diversify} are statistically rare, yet they dominate
real world risk and decision making \cite{STPDN,liu2024time}.
Modern forecasting models are typically trained by standard average risk
objectives on mixtures of regular and shifted data
\cite{patchtst,itransformer,ModernTCN,Sundial,kudrat2025patch}.
In such mixed ID/OOD distributions, frequent ID patterns dominate
optimization and representation learning, while rare OOD signals receive
limited influence.
Temporal embeddings, spatial embeddings, and learned graph structures
\cite{wavenet,agcrn,AdpSTGCN} can further reinforce high frequency regularities.
As a result, a model may remain accurate under average case benchmark
metrics while failing during rare, high impact deviations whose signals
have limited influence on representation learning
\cite{yichang, TimeStacker,parkwill}
(Fig.~\ref{fig:overview}(b)).

Real world dynamical systems still contain strong reusable structure.
Periodic rhythms, long term trends, and coordinated interactions across
variables are highly reusable across time and conditions \cite{STID},
while OOD events typically appear as deviations from this
shared latent subspace. These events often modify the stable process
through localized residual changes.
Existing decomposition, low rank, and dual branch forecasting methods
\cite{Autoformer,dlinear,Memformer,DualCast,bubianxing} exploit related
structure, usually to improve average case prediction in observation space.
This leaves a gap between using structural priors for average case accuracy
and designing a forecasting model around rare OOD signal suppression,
stable latent structure, residual deviations, and deployable latent inference.

These observations suggest a design principle for OOD robust forecasting.
Stable dynamics should be represented by a compact shared structure, while
rare deviations should be modeled outside the stable representation. Guided
by the theoretical principles, we propose \textbf{VLBM}, which shifts time
series prediction from observation space into a latent space. First, VLBM
learns a capacity controlled shared latent basis whose span defines a
low rank subspace for stable in distribution dynamics. Second, it explicitly
decomposes historical representations into basis subspace and orthogonal
residual components, and uses an Orthogonal Base--Residual generator to route
stable dynamics and OOD induced deviations to separate paths. Third, it aligns
a \emph{future aware posterior} with a \emph{future blind prior} over basis
activations, distilling future informed latent states into a test time
inference model that depends only on historical input. Our main contributions
are as follows.
\begin{itemize}
\item We identify and formulate an overlooked reliability problem in multivariate time series forecasting: under mixed ID/OOD conditions, rare but high impact event sequences are often diluted by average case training and evaluation, although they are critical to real world forecasting reliability.

\item We formalize rare OOD suppression under standard mixture risk training and derive a structured excess risk bound that links test time risk to stable low rank approximation, orthogonal residual approximation, and the posterior prior inference gap.

\item We propose \textbf{VLBM}, a variational latent basis forecasting framework that realizes this risk view through sample specific latent basis activation, Orthogonal Base--Residual generation, and posterior prior transfer from a future aware posterior to a future blind prior.

\item We construct real world OOD traffic forecasting benchmarks by
aligning traffic measurements with incident and lane closure records, and
demonstrate state of the art performance across ID benchmarks, OOD benchmarks,
and a synthetic simulation dataset.
\end{itemize}

\section{Related Work}

\subsection{Multivariate time series forecasting}

Multivariate time series forecasting serves as a fundamental scientific application with broad impact across multiple domains. The rapid advancement in this field is largely propelled by Transformer based architectures. To capture long range dependencies in extensive sequences, recent works have augmented the standard self attention mechanism with tailored structural priors. Notable innovations include trend seasonal decomposition \cite{Autoformer}, frequency domain mapping \cite{fredeformer}, channel aware representation learning \cite{crossformer, itransformer}, and specialized attention variants \cite{DGraFormer}. As model capacity and data scale continue to grow, this research direction has further evolved into LLM based solutions \cite{Time-LLM, TimeCMA} and general purpose Time Series Foundation Models \cite{Moria, Time-MOE, ROSE}, which aim to transfer large scale temporal knowledge across heterogeneous datasets and downstream forecasting tasks. In parallel, another line of research focuses on improving forecasting efficiency through lightweight architectures. MLP based models \cite{dlinear, timemixer, timebase, tqnet} and CNN based models \cite{TimesNet, ModernTCN} have demonstrated that carefully designed simple predictors can achieve competitive performance while reducing computational overhead. Despite their computational gains, existing designs often struggle to model explicit inter variable dependencies, as they primarily focus on isolated temporal extraction or implicit channel mixing. This limitation becomes more pronounced in complex multivariate systems, where variables are often coupled through latent physical, spatial, or functional relationships. Adaptive graph based forecasting methods have been proposed to explicitly infer and exploit inter variable relationships \cite{Memformer, MSGNet, TimeFilter,agga-mvfln,Lei2026}. These GNN based approaches construct feature graphs or dynamic relational structures, enabling models to capture dependency patterns that are difficult to represent using purely temporal operators. Meanwhile, emerging architectures such as KAN \cite{TIMEKAN} and Mamba \cite{timepro} are also being explored for time series forecasting, offering new perspectives on nonlinear function approximation and efficient long term modeling. 

\subsection{Out of Distribution forecasting}

Out of distribution (OOD) scenarios occur when testing environments deviate from the data distribution observed during training \cite{yizhixing}. Such shifts are common in real world time series, where regime changes, abnormal events, sensor drift, and external disturbances can invalidate the standard assumption that training and testing data are identically distributed \cite{Stone}. As a result, trained models may suffer substantial degradation when deployed under unseen conditions.

To improve robustness, recent forecasting studies have increasingly adopted decoupled or dual branch architectures. For example, graph based spatio temporal models separate heterogeneous traffic dynamics to reduce shift induced interference \cite{D2STGNN,STPDN}, while dual stream forecasting designs further disentangle intrinsic temporal patterns from irregular or contextual factors \cite{DualCast,BiST}. Related work also explores latent basis representations for modeling hidden disturbance factors in traffic prediction \cite{STEVE}. Many of these approaches are built upon domain specific assumptions, such as traffic topology, event driven disturbances, or handcrafted contextual structures, which limits their direct extension to general multivariate OOD forecasting. Another relevant line of work is time series anomaly detection, which provides useful tools for recognizing abnormal patterns and distributional irregularities. Frequency aware and channel aware methods are effective for identifying anomalous subsequences \cite{CATCH}, while recent benchmarks and adaptive anomaly detection frameworks further broaden the evaluation and modeling of abnormal temporal behaviors \cite{TAB,DADA}. 

Anomaly detection methods are typically optimized for detection, reconstruction, or anomaly scoring, with future prediction under shifted distributions receiving less direct attention. This leaves a gap between identifying abnormal observations and forecasting their subsequent effects. Motivated by this gap, we consider a forecasting oriented OOD formulation that explicitly separates stable dynamics from predictive deviations, aiming to provide a more general framework for multivariate forecasting under distribution shift.

\section{Theoretical Principles for OOD Robust Forecasting}
\label{sec:theory}

Before presenting VLBM, we derive the design principles required by OOD robust forecasting. We consider multivariate time series forecasting under distribution shift, where \(X_{\mathrm{raw}}\in\mathbb R^{T_{\mathrm{in}}\times N}\) denotes the historical input window and \(Y_{\mathrm{raw}}\in\mathbb R^{T_{\mathrm{out}}\times N}\) denotes the future sequence. The goal is to learn a predictor \(\hat f:X_{\mathrm{raw}}\mapsto \hat Y_{\mathrm{raw}}\) that remains accurate under both regular in distribution (ID) conditions and rare but high impact out of distribution (OOD) event intervals that are critical to forecasting reliability.

Real world time series often arise from a mixture of regular ID patterns and rare OOD events. We model the data distribution as
\begin{equation}
\mathcal D_\rho
=
(1-\rho)\mathcal D_{\mathrm{id}}
+
\rho\mathcal D_{\mathrm{ood}},
\qquad
0<\rho\ll 1,
\label{eq:mixture-distribution}
\end{equation}
where \(\mathcal D_{\mathrm{id}}\) corresponds to regular ID dynamics and \(\mathcal D_{\mathrm{ood}}\) corresponds to shifted or abnormal OOD conditions. After embedding, we denote the representations of \(X_{\mathrm{raw}}\) and \(Y_{\mathrm{raw}}\) by \(X,Y\in\mathbb R^{N\times d}\), where \(d\) is the embedding dimension. All projections in this section are applied along the embedding dimension.

The key question is how to design a predictor that preserves regular dynamics while remaining responsive to rare, reliability critical deviations. We answer this question through three linked principles. First, standard mixture risk suppresses rare OOD gradients, which motivates separating stable dynamics from residual deviations. Second, stable dynamics can be represented by a shared low rank basis whose span acts as a reusable library of regular patterns, and deviations outside this span are routed to an orthogonal residual channel. Third, once such a basis is learned, forecasting still requires sample specific basis activations. Future observations reveal useful activation information during training but are unavailable at test time, which motivates posterior prior transfer from a future aware posterior to a future blind prior.

\subsection{Rare OOD Suppression Under Mixture Risk Training}
\label{subsec:rare-ood-suppression}

For any predictor \(f\), define the mixture risk as
\begin{equation}
\mathcal R(f)
=
(1-\rho)\mathcal R_{\mathrm{id}}(f)
+
\rho\mathcal R_{\mathrm{ood}}(f).
\label{eq:mixture-risk}
\end{equation}
Here, \(\mathcal R_{\mathrm{id}}(f)=\mathbb E_{\mathcal D_{\mathrm{id}}}[\|Y-f(X)\|^2]\) denotes the risk under regular ID dynamics, and \(\mathcal R_{\mathrm{ood}}(f)=\mathbb E_{\mathcal D_{\mathrm{ood}}}[\|Y-f(X)\|^2]\) denotes the risk under shifted or abnormal OOD conditions.

\noindent\textbf{Lemma 1 (Gradient suppression under rare OOD mixtures).}
Assume that \(\mathcal R_{\mathrm{id}}\) and \(\mathcal R_{\mathrm{ood}}\) are differentiable. If there exists \(G>0\) such that \(\|\nabla \mathcal R_{\mathrm{ood}}(f)-\nabla \mathcal R_{\mathrm{id}}(f)\|\le G\), then
\begin{equation}
\left\|
\nabla \mathcal R(f)
-
\nabla \mathcal R_{\mathrm{id}}(f)
\right\|
\le
\rho G.
\label{eq:grad-suppression}
\end{equation}

The proof is provided in the supplementary material. Lemma~1 shows that the optimization signal induced by rare OOD events is suppressed at order \(O(\rho)\) under standard mixture risk training. When \(\rho\) is small, representation learning is therefore dominated by ID dynamics. This motivates an explicit deviation path for rare OOD signals. To formalize this mechanism, we use the Bayes predictor \(f^\star(X)=\mathbb E[Y\mid X]\) under the mixed distribution \(\mathcal D_\rho\) as the population level reference target, and assume that it admits a stable and residual decomposition.

\noindent\textbf{Assumption 1 (Latent ID and OOD decomposition).}
There exists a stable ID subspace \(\mathcal S^\star\subset\mathbb R^d\), with projection \(P_\star\in\mathbb R^{d\times d}\) and \(P_\star^\perp=I-P_\star\), such that
\begin{equation}
f^\star(X)
=
f_{\mathrm{id}}^\star(XP_\star)
+
f_{\mathrm{ood}}^\star(XP_\star^\perp).
\label{eq:latent-decomp}
\end{equation}
The first term represents stable and reusable ID dynamics, and the second term represents deviations outside the stable subspace. We further assume that the residual component contains no predictable mean once the ID component is given.
\begin{equation}
\mathbb E
\left[
f_{\mathrm{ood}}^\star(XP_\star^\perp)
\mid
XP_\star
\right]
=
0.
\label{eq:conditional-centered}
\end{equation}
Assumption~1 should be read as an approximate structural condition, with no requirement that it be an exact generative claim. It is most plausible when recurrent seasonal patterns, long term trends, and cross variable interactions can be captured by a stable low dimensional representation, while rare events mainly appear as residual deviations around that representation. If the ID dynamics and OOD deviations are strongly coupled, part of the deviation may be absorbed into the stable basis, which can weaken the separation between the base and residual paths.

This decomposition gives the first architectural principle. Stable dynamics and residual deviations should be represented by different channels. VLBM uses a Base--Residual generator, where the base path targets \(f_{\mathrm{id}}^\star\) and the residual path targets \(f_{\mathrm{ood}}^\star\).

\subsection{Low Rank Latent Basis and Orthogonal Residual Channel}
\label{subsec:low-rank-residual}

We next derive why the stable channel should be low rank and why the residual channel should receive an orthogonal component. Stable dynamics often recur across samples and variables, even when rare deviations occur. Let
\(\mathcal M_{\mathrm{id}}=\{f_{\mathrm{id}}^\star(XP_\star):X\sim\mathcal D_\rho\}\)
denote the family of stable ID components induced by the mixed distribution. We quantify its low dimensional approximability by the Kolmogorov \(M\) width~\cite{pinkus2012n},

\begin{equation}
d_M(\mathcal M_{\mathrm{id}})
=
\inf_{\dim(\mathcal S)=M}
\sup_{u\in\mathcal M_{\mathrm{id}}}
\inf_{v\in\mathcal S}
\|u-v\|_{L^2},
\label{eq:kolmogorov-width}
\end{equation}
where \(\mathcal S\) denotes a candidate \(M\) dimensional linear subspace used to approximate the stable dynamics family \(\mathcal M_{\mathrm{id}}\), and \(M\) corresponds to the target latent dimension. Intuitively, \(d_M(\mathcal M_{\mathrm{id}})\) is the smallest worst case error for approximating all stable ID components using an \(M\) dimensional subspace.

Let \(\mathcal S_B\) denote a learned \(M\) dimensional stable subspace, and let \(P_B\) be the projection onto \(\mathcal S_B\), with \(P_B^\perp=I-P_B\). We write \(X_{\parallel}^{(B)}=XP_B\) and \(X_{\perp}^{(B)}=XP_B^\perp\). The term \(X_{\parallel}^{(B)}\) is the stable projection of the historical representation, and \(X_{\perp}^{(B)}\) captures the component outside the learned stable subspace.

Define the ID approximation error as
\begin{equation}
\mathcal A_{\mathrm{id}}(B)
=
\inf_g
\mathbb E
\left\|
g(XP_B)
-
f_{\mathrm{id}}^\star(XP_\star)
\right\|^2.
\label{eq:id-approx-error}
\end{equation}
Under the approximation capacity condition of the base channel, if the learned subspace \(\mathcal S_B\) associated with \(P_B\) is a \(C_B\) approximate minimizer of the Kolmogorov width problem, then
\begin{equation}
\mathcal A_{\mathrm{id}}(B)
\le
C_B^2 d_M^2(\mathcal M_{\mathrm{id}}).
\label{eq:id-width-bound}
\end{equation}
The derivation is provided in the supplementary material. This compact result supports the use of a shared low rank latent basis \(B\) for stable ID dynamics. It also motivates controlling the basis size \(M\). The value of \(M\) should be large enough to approximate recurring stable dynamics, while remaining finite enough to preserve separation between stable structure and distribution specific deviations.

The same low rank restriction also creates a limitation. Components outside the stable subspace remain invisible to the base channel. Consider a base only predictor \(f_{\mathrm{base}}(X)=g(XP_\star)\), which only depends on the ID projection.

\noindent\textbf{Proposition 1 (Irreducible OOD bias of base only predictors).}
Under Assumption~1,
\begin{equation}
\begin{aligned}
\inf_g
\left[
\mathcal R(g(XP_\star))-\mathcal R(f^\star)
\right]
&=
\mathbb E
\left\|
f_{\mathrm{ood}}^\star(XP_\star^\perp)
\right\|^2.
\end{aligned}
\label{eq:base-only-lower}
\end{equation}

The proof is provided in the supplementary material. Proposition~1 shows that any predictor restricted to the ID projection leaves an irreducible residual error whenever the OOD component has nonzero energy. This explains why a base only stable subspace is insufficient, and why the low rank stable channel should be paired with a complementary residual channel.

We measure the approximation ability of this residual channel by
\begin{equation}
\mathcal A_{\perp}(B)
=
\inf_r
\mathbb E
\left\|
r(X_{\perp}^{(B)},X)
-
f_{\mathrm{ood}}^\star(XP_\star^\perp)
\right\|^2.
\label{eq:residual-approx-error}
\end{equation}
Here, \(X_{\perp}^{(B)}\) denotes the component orthogonal to the learned stable subspace, and \(X\) provides contextual conditioning for interpreting that deviation. Thus, \(\mathcal A_{\perp}(B)\) measures how well the residual channel approximates the oracle OOD component when the explicit deviation signal is anchored by the current historical context. VLBM constructs \(P_B\) from the learned basis, uses \(X_{\parallel}^{(B)}=XP_B\) in the base path, and routes \(X_{\perp}^{(B)}=X(I-P_B)\) to the residual path.

\subsection{Posterior Prior Transfer for Test Time Inference}
\label{subsec:oracle-distillation}

The previous principles specify how stable and residual information should be routed. They also leave one inference problem. Once a shared stable basis \(B\) is learned, the model must determine which basis patterns are active for the current sample. Given the learned basis \(B\) introduced in Section~\ref{subsec:low-rank-residual}, whose row span defines \(\mathcal S_B\), we introduce a sample specific basis activation \(\mathbf w\) and write \(z_B(\mathbf w)=B^\top\mathbf w\in\mathcal S_B\). The projection \(X_{\parallel}^{(B)}\) is the historical representation projected into the learned basis library, and \(z_B(\mathbf w)\) is the activated latent state used for prediction. Thus, \(P_B\) defines the shared subspace geometry, \(X_{\parallel}^{(B)}\) provides projected historical evidence, and \(\mathbf w\) determines the sample specific activation of basis patterns. The orthogonal component \(X_{\perp}^{(B)}\) is passed to the residual channel so rare deviations keep a dedicated representation outside the stable basis. Let \(\Gamma_B(X,\mathbf w)=(z_B(\mathbf w),X_{\parallel}^{(B)},X_{\perp}^{(B)})\) collect these conditioning variables, where \(B\) is treated as a model parameter in the conditional predictive model.

At the population level, this gives the following latent predictive formulation.
\begin{equation}
p_{\theta,\psi,B}(Y\mid X)
=
\int
p_{\theta,B}
\left(
Y\mid \Gamma_B(X,\mathbf w)
\right)
p_{\psi,B}(\mathbf w\mid X)
d\mathbf w.
\label{eq:latent-marginal}
\end{equation}
Here, \(p_{\psi,B}(\mathbf w\mid X)\) is a future blind prior that infers the latent activation from the historical input \(X\) under the learned basis, and \(p_{\theta,B}(Y\mid\Gamma_B(X,\mathbf w))\) models the future conditioned on the stable basis induced pattern, the projected stable context, and the orthogonal residual component.

During training, the future sequence \(Y\) is observed and provides information about which latent activation \(\mathbf w\) best explains the realized future. We therefore introduce a future aware posterior \(q_{\phi,B}(\mathbf w\mid X,Y)\). For any such posterior, Jensen's inequality yields the evidence lower bound
\begin{equation}
\begin{aligned}
\log p_{\theta,\psi,B}(Y\mid X)
\ge\;&
\mathbb E_{\mathbf w\sim q_{\phi,B}(\cdot\mid X,Y)}
\left[
\log p_{\theta,B}
\left(
Y\mid\Gamma_B(X,\mathbf w)
\right)
\right]\\
&-
\mathrm{KL}
\left(
q_{\phi,B}(\mathbf w\mid X,Y)
\|p_{\psi,B}(\mathbf w\mid X)
\right).
\end{aligned}
\label{eq:theory-elbo}
\end{equation}
The first term requires the stable pattern \(z_B(\mathbf w)\), the projected stable context \(X_{\parallel}^{(B)}\), and the orthogonal residual component \(X_{\perp}^{(B)}\) to jointly explain the future sequence. The second term aligns the future aware posterior with the future blind prior. This KL term supports posterior prior transfer and also acts as an information bottleneck. The prior \(p_{\psi,B}(\mathbf w\mid X)\) conditions only on \(X\), so posterior information about the realized future that lacks support in \(X\) incurs a KL cost. The variational objective therefore encourages \(\mathbf w\) to encode history predictable basis activations. Unpredictable deviations are handled by the orthogonal residual channel.

Let \(G_{\theta,B}(X,\mathbf w)\) denote the population level predictor induced by the conditional model in Eq.~\eqref{eq:latent-marginal}. Define the posterior induced predictor
\begin{equation}
F_q^B(X,Y)
=
\mathbb E_{\mathbf w\sim q_{\phi,B}(\cdot\mid X,Y)}
G_{\theta,B}(X,\mathbf w),
\end{equation}
and the prior induced predictor
\begin{equation}
F_p^B(X)
=
\mathbb E_{\mathbf w\sim p_{\psi,B}(\cdot\mid X)}
G_{\theta,B}(X,\mathbf w).
\label{eq:posterior-prior-predictors}
\end{equation}

\noindent\textbf{Proposition 2 (Posterior prior transfer controls test time inference gap).}
Assume that \(G_{\theta,B}(X,\mathbf w)\) is \(L_G\)-Lipschitz in \(\mathbf w\), i.e., \(\|G_{\theta,B}(X,\mathbf w)-G_{\theta,B}(X,\mathbf w')\|\le L_G\|\mathbf w-\mathbf w'\|\) for fixed \(X\), and that the prior satisfies a Talagrand \(T_2(c)\) transport inequality, i.e., \(W_2^2(q,p_{\psi,B})\le 2c\,\mathrm{KL}(q\|p_{\psi,B})\) for any admissible \(q\)~\cite{otto2000generalization,villani2009optimal}. Then
\begin{equation}
\begin{aligned}
&\mathbb E
\|F_q^B(X,Y)-F_p^B(X)\|^2\\
&\quad \le
2cL_G^2
\mathbb E
\mathrm{KL}
\bigl(
q_{\phi,B}(\mathbf w\mid X,Y)
\|p_{\psi,B}(\mathbf w\mid X)
\bigr).
\end{aligned}
\label{eq:kl-gap-bound}
\end{equation}

The proof is provided in the supplementary material. The \(T_2(c)\) condition is a standard regularity assumption for distributions with controlled concentration. For the learned Gaussian prior used by VLBM, it can be interpreted as requiring a bounded variance scale, with \(c\) reflecting an upper bound on the prior dispersion. Proposition~2 shows that posterior prior transfer controls the gap between the training time posterior predictor and the test time prior predictor under the learned basis \(B\). VLBM trains a future aware posterior \(q_{\phi,B}(\mathbf w\mid X,Y)\) as a teacher and aligns it with a future blind prior \(p_{\psi,B}(\mathbf w\mid X)\), which is the only latent inference model used at test time.

\subsection{Structured Risk Summary}
\label{subsec:structured-risk-summary}

We now summarize the theoretical chain by combining the three sources of error, namely the low dimensional approximation error of stable ID dynamics, the orthogonal residual approximation error, and the posterior prior inference gap controlled by the KL term. Since only the prior induced predictor is available at deployment, the final summary is stated for \(F_p^B\), with \(\mathcal R(F_p^B)=\mathbb E_{\mathcal D_\rho}\|Y-F_p^B(X)\|^2\).

\noindent\textbf{Theorem 1 (Structured excess risk bound).}
Under Assumption~1, the regularity conditions in Proposition~2, and the approximation conditions of the base and residual generator, if the learned stable subspace associated with \(P_B\) is a \(C_B\) approximate minimizer of the Kolmogorov width problem, then there exist constants \(C_{\mathrm{id}}, C_{\perp}, C_{\mathrm{kl}}>0\) such that the test time prior predictor satisfies
\begin{equation}
\begin{aligned}
&\mathcal R(F_p^B)-\mathcal R(f^\star)\\
&\quad \le
C_{\mathrm{id}} C_B^2 d_M^2(\mathcal M_{\mathrm{id}})
+
C_{\perp}\mathcal A_{\perp}(B)\\
&\qquad+
C_{\mathrm{kl}} cL_G^2
\mathbb E
\mathrm{KL}
\bigl(
q_{\phi,B}(\mathbf w\mid X,Y)
\|p_{\psi,B}(\mathbf w\mid X)
\bigr).
\end{aligned}
\label{eq:oracle-width}
\end{equation}

The proof is provided in the supplementary material. Theorem~1 links the three error terms to the three central modules of VLBM. The first term motivates the shared low rank latent basis for stable ID dynamics. The second term motivates the orthogonal residual channel for deviations outside the basis subspace. In practice, these two approximation terms are optimized through the supervised forecasting loss on the Base--Residual prediction. The third term is controlled directly by the posterior prior KL penalty.

%%%%%%%%%%%%%%%%%%%%%%%%%%%%%%%%%%%%%%%%%%%%%%%%%%%%%%%%%%%%%%%%%%%%%%%%%%%%%%%
\begin{figure}[t]
\centering
\includegraphics[width=0.85\linewidth]{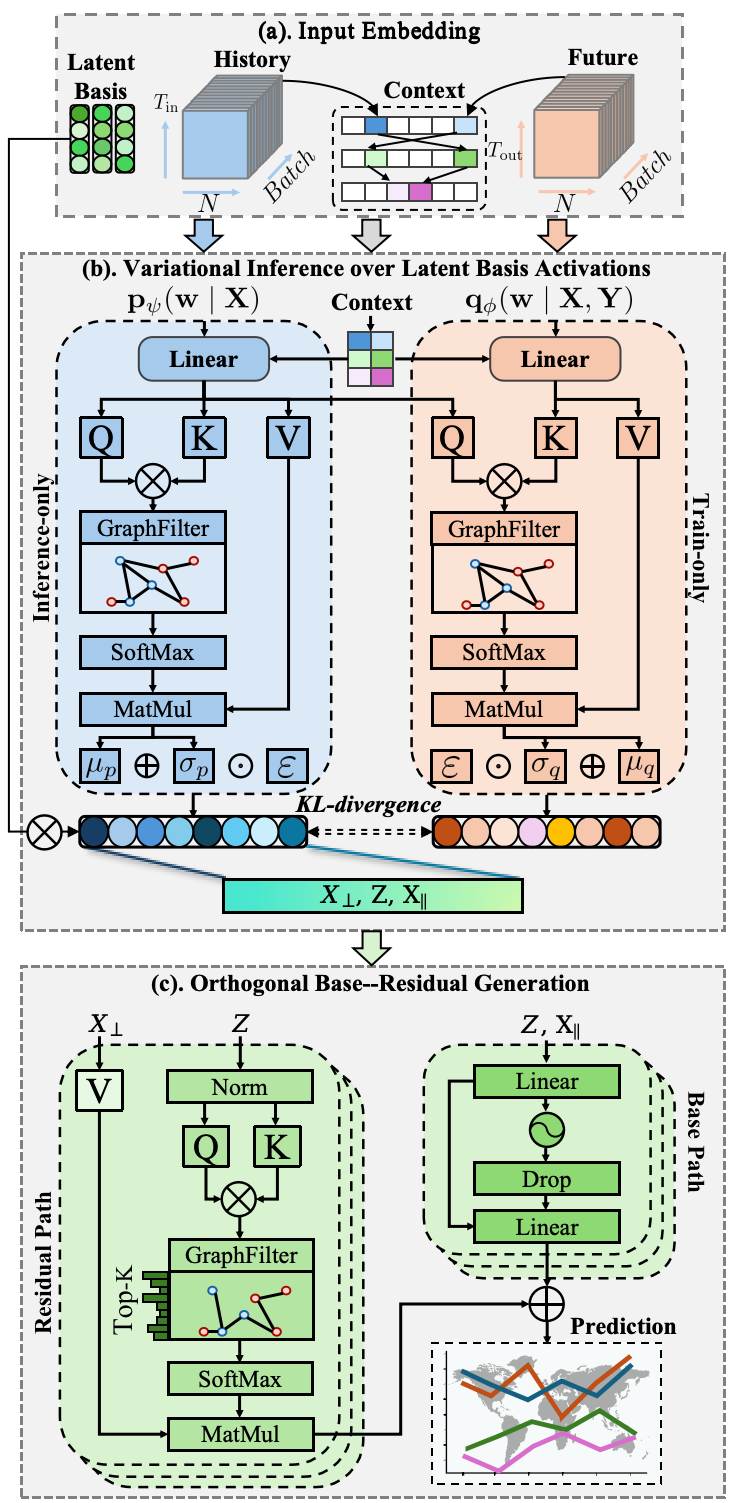}
\caption{
Overall architecture of VLBM.
(a) Input embedding maps raw time series into variable-level representations.
(b) Variational inference over latent basis activations aligns a future-aware posterior with a future-blind prior.
(c) Orthogonal Base--Residual generation separates stable dynamics from deviation components.
}
\label{fig:fig2}
\end{figure}
\section{The VLBM Framework}
\label{sec:method}

Guided by Section~\ref{sec:theory}, VLBM implements OOD robust forecasting through three central designs that correspond to the main terms in the structured risk summary: a shared low rank basis for stable approximation, an explicit orthogonal residual channel for deviation approximation, and posterior prior transfer for controlling the test time inference gap. After a common input embedding, the basis induced variational module aligns a future aware posterior \(q_\phi(\mathbf w\mid X,Y)\) with a future blind prior \(p_\psi(\mathbf w\mid X)\), and the Orthogonal Base--Residual generator uses the learned basis \(B\) to construct \(P_B\), decompose \(X\) into \(X_{\parallel}^{(B)}\) and \(X_{\perp}^{(B)}\), and route the two components to the base and residual paths. In this design, \(P_B\) defines the shared geometric decomposition, and the latent weights \(\mathbf w\) specify sample specific activations over the basis.

\subsection{Input Embedding}
\label{subsec:embedding}

Given a raw historical input window \(X_{\mathrm{raw}}\in\mathbb R^{N_b\times T_{\mathrm{in}}\times N}\), where \(N_b\) is the batch size, \(T_{\mathrm{in}}\) is the input length, and \(N\) is the number of variables, VLBM first maps it into a variable level representation \(X\in\mathbb R^{N_b\times N\times d}\). The embedding dimension \(d\) is consistent with Section~\ref{sec:theory}. This embedding compresses the temporal dimension into feature channels, so subsequent latent inference and graph based generation operate over variables.

The input embedding consists of four components, namely value projection, time of day embedding, day of week embedding, and variable identity embedding. The value channel is projected along the temporal dimension by convolution, producing \(E_{\mathrm{val}}\in\mathbb R^{N_b\times N\times d}\). The time of day and day of week indices at the last input step are embedded as \(E_{\mathrm{time}}\in\mathbb R^{N_b\times N\times d}\) and \(E_{\mathrm{day}}\in\mathbb R^{N_b\times N\times d}\). Variable identity is represented by a learnable node embedding \(E_{\mathrm{node}}\in\mathbb R^{N_b\times N\times d}\). The final historical representation is obtained by concatenating the four embeddings and projecting them back to dimension \(d\).
\begin{equation}
X
=
\mathrm{Conv}_{\mathrm{emb}}
\left(
\mathrm{Concat}
[
E_{\mathrm{time}},
E_{\mathrm{day}},
E_{\mathrm{val}},
E_{\mathrm{node}}
]
\right).
\label{eq:input-embedding}
\end{equation}
During training, the future sequence \(Y_{\mathrm{raw}}\) is embedded using the same form to obtain \(Y\in\mathbb R^{N_b\times N\times d}\), which is used only for constructing the future aware posterior. During inference, only the historical representation \(X\) is available.

\subsection{Basis Induced Variational Inference}
\label{subsec:vi}

VLBM uses a shared latent basis to instantiate the stable latent subspace introduced in Section~\ref{sec:theory}. Let
\begin{equation}
B\in\mathbb R^{M\times d}
\end{equation}
denote the learnable basis, where \(M\) is the number of basis vectors and controls the capacity of the stable subspace. The row span of \(B\) defines the learned stable subspace \(\mathcal S_B\). For each variable \(n\), VLBM infers a latent weight vector \(w_n\in\mathbb R^M\), and induces a stable latent state through
\begin{equation}
z_n
=
B^\top w_n
\in
\mathbb R^d.
\label{eq:latent-state}
\end{equation}
Stacking all variables yields \(\mathbf Z\in\mathbb R^{N_b\times N\times d}\). This corresponds to \(z_B(\mathbf w)\) in Section~\ref{sec:theory}. The shared basis \(B\) provides reusable stable patterns, while the latent weights \(\mathbf w\) specify their activations for each sample. The posterior and prior share \(B\), so training and inference use the same latent coordinate system.

\subsubsection{Future Aware Posterior and Future Blind Prior}

During training, the future representation \(Y\) is available and provides information about which latent activation explains the realized future. Therefore, VLBM uses a future aware posterior \(q_\phi(\mathbf w\mid X,Y)\). During inference, \(Y\) is unavailable, so VLBM relies on a future blind prior \(p_\psi(\mathbf w\mid X)\), which infers latent activations from the historical representation only.

Both distributions are parameterized as variable wise Gaussian distributions.
\begin{equation}
\begin{aligned}
(\mu_q,\log\sigma_q^2)&=\Phi_q(X,Y),\\
(\mu_p,\log\sigma_p^2)&=\Phi_p(X),
\end{aligned}
\label{eq:posterior-prior-params}
\end{equation}
where \(\mu_q,\mu_p,\log\sigma_q^2,\log\sigma_p^2\in\mathbb R^{N_b\times N\times M}\). The posterior provides future aware latent activations during training, and the prior provides latent activations available at test time. The alignment between them transfers future aware latent inference into the future blind prior.

\subsubsection{Graph Structured Encoders}

The posterior encoder \(\Phi_q\) and the prior encoder \(\Phi_p\) use the same graph structured aggregation form with different inputs. Let \(U\) denote the query side input and \(V\) denote the value side input. For the posterior encoder, \((U,V)=(X,Y)\). For the prior encoder, \((U,V)=(X,X)\). The two inputs are first projected by \(1\times1\) convolutions with nonlinear activation.
\begin{equation}
\tilde U=\sigma(\mathrm{Conv}_{u}(U)),
\qquad
\tilde V=\sigma(\mathrm{Conv}_{v}(V)).
\end{equation}
For variables \(i\) and \(j\), the attention score is computed by
\begin{equation}
e_{ij}
=
\mathrm{LeakyReLU}
\left(
a^\top
[
\tilde u_i,\tilde v_j
]
\right),
\label{eq:encoder-attn-score}
\end{equation}
where \([\cdot,\cdot]\) denotes concatenation. Given an adjacency matrix \(A\), invalid edges are masked before normalization by applying \(\alpha_{ij}=\mathrm{softmax}_{j}(e_{ij}+m_{ij})\), with \(m_{ij}=0\) for \(A_{ij}>0\) and \(m_{ij}=-\infty\) otherwise. The variable level representation is obtained by attention weighted aggregation, \(h_i=\sum_{j=1}^{N}\alpha_{ij}\tilde v_j\).
For datasets with a known physical topology, \(A\) uses that topology as a structural prior. For datasets without such information, \(A\) is set to the identity mask.
Two linear heads then output the Gaussian parameters of the latent weights.
\begin{equation}
\mu_i=W_\mu h_i+b_\mu,
\qquad
\log\sigma_i^2=W_\sigma h_i+b_\sigma.
\label{eq:gaussian-heads}
\end{equation}
The posterior branch uses cross interaction between history and future, allowing the latent weights to reflect which basis patterns explain the realized future. The prior branch uses only historical self interaction, so the same type of activation must be inferred from deployable historical information.

\subsubsection{Reparameterized Sampling}

During training, latent weights are sampled from the posterior by the reparameterization trick.
\begin{equation}
w_n^{(q)}
=
\mu_{q,n}
+
\sigma_{q,n}\odot\varepsilon_n,
\qquad
\varepsilon_n\sim\mathcal N(0,I).
\label{eq:posterior-reparam}
\end{equation}
During inference, latent weights are sampled from the prior.
\begin{equation}
w_n^{(p)}
=
\mu_{p,n}
+
\sigma_{p,n}\odot\varepsilon_n,
\qquad
\varepsilon_n\sim\mathcal N(0,I).
\label{eq:prior-reparam}
\end{equation}
The corresponding latent states are \(z_n^{(q)}=B^\top w_n^{(q)}\) and \(z_n^{(p)}=B^\top w_n^{(p)}\). The generator is driven by prior induced latent states \(\mathbf Z^{(p)}\), so the prediction path is consistent between training and inference. The posterior contributes through KL alignment, acting as a future aware teacher for the prior.

\subsection{Orthogonal Base--Residual Generation}
\label{subsec:gen}

After obtaining the prior induced latent states \(\mathbf Z^{(p)}\), VLBM generates the future prediction through an Orthogonal Base--Residual generator. This generator explicitly implements the stable and residual decomposition in Section~\ref{sec:theory}. The shared basis \(B\) defines the stable subspace, and the input representation \(X\) is projected onto this subspace and its orthogonal complement.

Given \(B\in\mathbb R^{M\times d}\), VLBM constructs a ridge stabilized projection matrix
\begin{equation}
P_B
=
B^\top(BB^\top+\eta I_M)^{-1}B,
\label{eq:projection-matrix}
\end{equation}
where \(\eta>0\) is a small numerical ridge that stabilizes the projection when \(BB^\top\) is ill conditioned. When \(B\) has full row rank and \(\eta\to0\), \(P_B\) becomes the orthogonal projection onto \(\mathcal S_B=\mathrm{span}(B)\). The input representation is decomposed as
\begin{equation}
X_{\parallel}^{(B)}
=
XP_B,
\qquad
X_{\perp}^{(B)}
=
X(I-P_B).
\label{eq:orthogonal-decomposition}
\end{equation}
Here, \(X_{\parallel}^{(B)}\) is the component inside the stable subspace, and \(X_{\perp}^{(B)}\) is the component orthogonal to that subspace.
The projection \(P_B\) and the latent state \(\mathbf Z^{(p)}\) play different roles. \(P_B\) defines the shared geometric decomposition induced by \(B\). The latent state \(\mathbf Z^{(p)}\) is the prior inferred, sample specific stable state used by the base path.

\subsubsection{Base Path for Stable Prediction Within the Latent Subspace}

The base path models stable and reusable dynamics. It takes the prior induced latent states \(\mathbf Z^{(p)}\) and the projected stable component \(X_{\parallel}^{(B)}\) as input.
\begin{equation}
P_0
=
\mathrm{Norm}_{b}
\left(
\mathrm{Concat}
[
\mathbf Z^{(p)},
X_{\parallel}^{(B)}
]
\right).
\label{eq:base-input}
\end{equation}
Here, \(\mathbf Z^{(p)}\) provides the prior inferred stable latent state for future prediction. The term \(X_{\parallel}^{(B)}\) provides the projected stable context from the observed history. The base stream is refined by stacked residual convolution blocks.
\begin{equation}
P_{\ell+1}
=
P_{\ell}
+
\mathrm{Conv}_2
\left(
\mathrm{Drop}
\left(
\mathrm{ReLU}
\left(
\mathrm{Conv}_1(P_{\ell})
\right)
\right)
\right).
\label{eq:base-block}
\end{equation}
The final base stream is mapped to the base prediction \(\hat Y_{\mathrm{base}}\). Since the base path is conditioned on \(\mathbf Z^{(p)}\) and \(X_{\parallel}^{(B)}\), it focuses on stable dynamics associated with the shared latent subspace.

\subsubsection{Residual Path for Orthogonal Deviation Modeling}

The residual path models deviations outside the stable subspace. It takes the orthogonal residual component \(X_{\perp}^{(B)}\) together with the historical representation \(X\).
\begin{equation}
R_0
=
\mathrm{Norm}_{r}
\left(
\mathrm{Concat}
[
X_{\perp}^{(B)},
X
]
\right).
\label{eq:residual-input}
\end{equation}
The term \(X_{\perp}^{(B)}\) isolates information unexplained by the shared latent basis, and \(X\) provides contextual anchoring for interpreting the deviation under the current temporal and variable state. This path gives OOD induced deviations a dedicated modeling channel after the stable subspace has been separated.

The residual path uses a lightweight latent induced graph to share deviation information among related variables. From the prior induced states \(\mathbf Z^{(p)}\), VLBM computes cosine similarities \(S_{ij}\), optionally fuses them with a prior topology \(A_{\mathrm{topo}}\), and applies TopK sparsification.
\begin{equation}
\begin{aligned}
S_{ij}
&=
\frac{
\langle z_i^{(p)},z_j^{(p)}\rangle
}{
\|z_i^{(p)}\|_2\|z_j^{(p)}\|_2
},\\
\tilde A^{(p)}
&=
(1-\alpha)S+\alpha A_{\mathrm{topo}},
\quad \alpha\in[0,1],\\
A^{(p)}
&=
\mathrm{Softmax}
\left(
\mathrm{TopK}
(
\tilde A^{(p)}
)
\right).
\end{aligned}
\label{eq:latent-induced-graph}
\end{equation}
This graph serves as a compact implementation device for residual propagation after the orthogonal decomposition. Since \(A^{(p)}\) is derived from the current prior induced states, it provides sample dependent routing in the stable latent coordinate system and can also incorporate optional topology information. Let \(R_\ell\) denote the residual representation after the \(\ell\)-th propagation block, with \(R_0\) initialized by Eq.~\eqref{eq:residual-input}. Each block first computes \(H_\ell=A^{(p)}R_\ell\), and then updates \(R_{\ell+1}\) through stacked residual convolution blocks. The final residual stream is mapped to the residual prediction \(\hat R_{\mathrm{res}}\).
The final forecast in the raw target space is formed by combining the base and residual predictions.
\begin{equation}
\hat Y_{\mathrm{raw}}
=
\hat Y_{\mathrm{base}}
+
\hat R_{\mathrm{res}}.
\label{eq:final-prediction}
\end{equation}
This design creates an OOD aware division of labor. The base path reconstructs stable dynamics in the learned latent subspace, and the residual path models orthogonal deviation signals.

\subsection{Objective and Inference}
\label{subsec:objective}

The objective follows the ELBO motivation in Eq.~\eqref{eq:theory-elbo}. The likelihood term is implemented as a supervised forecasting loss on the final Base--Residual prediction in the raw target space, and the KL term implements posterior prior transfer from \(q_\phi(\mathbf w\mid X,Y)\) to \(p_\psi(\mathbf w\mid X)\). The training objective is
\begin{equation}
\mathcal L_{\mathrm{train}}
=
\mathcal L_{\mathrm{rec}}
+
\beta\mathcal L_{\mathrm{KL}},
\label{eq:training-objective}
\end{equation}
where \(\beta>0\) controls the strength of posterior prior transfer. The reconstruction term corresponds to the negative log likelihood part of Eq.~\eqref{eq:theory-elbo}. In implementation, VLBM uses an \(\ell_1\) surrogate on the deployment aligned prior path.
\(
\mathcal L_{\mathrm{rec}}
=
\left\|
Y_{\mathrm{raw}}-
\left(
\hat Y_{\mathrm{base}}
+
\hat R_{\mathrm{res}}
\right)
\right\|_1.
\)
The consistency term is the KL divergence between the posterior and prior Gaussian distributions.
\(
\mathcal L_{\mathrm{KL}}
=
\mathrm{KL}
\left(
\mathcal N(\mu_q,\sigma_q^2)
\|
\mathcal N(\mu_p,\sigma_p^2)
\right).
\)

During training, the posterior uses the embedded pair \((X,Y)\) to provide future aware latent supervision, and the forecasting loss is supervised by the raw future sequence \(Y_{\mathrm{raw}}\). The KL term transfers posterior information into the future blind prior. The prediction path uses prior induced latent states \(\mathbf Z^{(p)}\) and the Orthogonal Base--Residual generator to produce \(\hat Y_{\mathrm{raw}}\), matching the information available at deployment. During inference, \(Y_{\mathrm{raw}}\) and its embedding \(Y\) are unavailable. VLBM uses only \(p_\psi(\mathbf w\mid X)\), constructs \(\mathbf Z^{(p)}\), and follows the same generator path to produce the forecast.

\begin{table}[!htbp]
\centering
\caption{Dataset Statistics in ID and OOD Scenarios}
\label{table-show}
\resizebox{\linewidth}{!}{% 
\begin{tabular}{@{}ccccc@{}}
\toprule
\textbf{Datasets} & \textbf{Time Series Length} & \textbf{Variables} & \textbf{Interval} & \textbf{Forecast Horizons} \\
\midrule
Weather & 52696 & 21 & 10 mins & [12, 96, 336, 720] \\
ECL & 26304 & 321 & 1 hr & [12, 96, 336, 720] \\
Solar & 35040 & 137 & 15 mins & [12, 96, 336, 720] \\
Flight & 26304 & 7 & 1 hr & [12, 96, 336, 720] \\
PEMS03 & 26208 & 358 & 5 mins & [12, 24, 48, 96] \\
PEMS04 & 16992 & 307 & 5 mins & [12, 24, 48, 96] \\
PEMS07 & 28224 & 883 & 5 mins & [12, 24, 48, 96] \\
PEMS08 & 17856 & 170 & 5 mins & [12, 24, 48, 96] \\
\midrule
MSL & 132046 & 55 & 1 min & [12, 48, 96, 192] \\
PSM & 220322 & 25 & 1 min & [12, 48, 96, 192] \\
CHP-LCS-Flow & 36288 & 365 & 5 mins & [12, 24, 48, 96] \\
CHP-LCS-Speed & 36288 & 365 & 5 mins & [12, 24, 48, 96] \\
\bottomrule
\end{tabular}
}
\end{table}

\begin{table*}[t]
\centering
\caption{Average forecasting performance under OOD evaluation, computed only on anomalous test intervals. Best results are highlighted in bold, and the second best results are underlined. The VLBM's reports represent the mean and standard deviation calculated from five randomly selected seeds}
\label{tab:ood}
\footnotesize
\setlength{\tabcolsep}{4.2pt}
\renewcommand{\arraystretch}{1.05}
% 如果超出页宽可解除下一行的注释
% \resizebox{\linewidth}{!}{
\begin{tabular}{@{}l cc cc cc cc c@{}}
\toprule
\multirow{2}{*}{\textbf{Models (Year)}} & \multicolumn{2}{c}{\textbf{MSL}} & \multicolumn{2}{c}{\textbf{PSM}} & \multicolumn{2}{c}{\textbf{CHP-LCS-Flow}} & \multicolumn{2}{c}{\textbf{CHP-LCS-Speed}} & \multirow{2}{*}{\textbf{1\textsuperscript{st} Count}} \\
\cmidrule(lr){2-3} \cmidrule(lr){4-5} \cmidrule(lr){6-7} \cmidrule(lr){8-9}
& MAE & MSE & MAE & MSE & MAE & MSE & MAE & MSE & \\
\midrule
DLinear (2023)       & 0.134 & 6.525 & 0.285 & 0.362 & 0.320 & 0.289 & 0.479 & 1.131 & 0 \\
MegaCRN (2023)       & \textbf{0.084} & 6.449 & 0.280 & 0.391 & 0.276 & 0.178 & 0.393 & 1.148 & \underline{1} \\
Memformer (2024)     & 0.124 & 7.228 & 0.266 & 0.361 & 0.260 & 0.156 & 0.441 & 1.218 & 0 \\
iTransformer (2024)  & 0.087 & 6.517 & 0.266 & 0.361 & 0.224 & 0.120 & 0.465 & 1.249 & 0 \\
TimeLLM (2024)       & 0.120 & 7.116 & 0.272 & 0.356 & 0.345 & 0.284 & 0.475 & 1.280 & 0 \\
ModernTCN (2024)     & 0.103 & 6.363 & 0.267 & \underline{0.354} & 0.233 & 0.149 & 0.420 & 1.250 & 0 \\
STONE (2024)         &  \underline{0.086} & 6.494 & \underline{0.260} & 0.355 & 0.212 & \underline{0.109} & 0.380 & 1.084 & 0 \\
TimeKAN (2025)       & \textbf{0.084} & 6.352 & 0.266 & 0.355 & 0.325 & 0.253 & 0.468 & 1.277 & \underline{1} \\
TimePro (2025)       & 0.115 & 6.696 & 0.265 & 0.361 & 0.253 & 0.148 & 0.380 & 1.094 & 0 \\
FilterTS (2025)      & 0.095 &  \underline{6.273} & 0.265 & \underline{0.354} & 0.281 & 0.183 & 0.448 & 1.283 & 0 \\
TimeMixer++ (2025)   & 0.118 & 6.424 & 0.265 & 0.361 & 0.222 & 0.128 & \underline{0.364} & \underline{1.052} & 0 \\
DUET (2025)          & 0.126 & 7.382 & 0.272 & 0.372 & \underline{0.190} & \textbf{0.094} & 0.580 & 1.619 & \underline{1} \\
\midrule
\multirow{2}{*}{\textbf{VLBM (Ours)}} 
& \textbf{0.084} & \textbf{6.129} & \textbf{0.243} & \textbf{0.332} & \textbf{0.155} & \underline{0.119} & \textbf{0.345} & \textbf{1.051} & \multirow{2}{*}{\textbf{7}} \\
& \color{gray}\scriptsize($\pm$0.0007) & \color{gray}\scriptsize($\pm$0.0201) & \color{gray}\scriptsize($\pm$0.0013) & \color{gray}\scriptsize($\pm$0.0016) & \color{gray}\scriptsize($\pm$0.0005) & \color{gray}\scriptsize($\pm$0.0018) & \color{gray}\scriptsize($\pm$0.0006) & \color{gray}\scriptsize($\pm$0.0127) & \\
\bottomrule
\end{tabular}
% }
\end{table*}

\begin{table*}[t]
\centering
\caption{Average forecasting performance under standard in-distribution (ID) evaluation. Best results are highlighted in bold, and the second best results are underlined. VLBM reports mean and standard deviation over five random seeds.}
\label{tab:id}
\scriptsize
\setlength{\tabcolsep}{2.2pt}
\renewcommand{\arraystretch}{1.05}
\begin{tabular}{@{}l cc cc cc cc cc cc cc cc c@{}}
\toprule
\multirow{2}{*}{\textbf{Models}} 
& \multicolumn{2}{c}{\textbf{PEMS03}} 
& \multicolumn{2}{c}{\textbf{PEMS04}} 
& \multicolumn{2}{c}{\textbf{PEMS07}} 
& \multicolumn{2}{c}{\textbf{PEMS08}} 
& \multicolumn{2}{c}{\textbf{Weather}} 
& \multicolumn{2}{c}{\textbf{ECL}} 
& \multicolumn{2}{c}{\textbf{Solar}} 
& \multicolumn{2}{c}{\textbf{Flight}} 
& \multirow{2}{*}{\textbf{\#1st}} \\
\cmidrule(lr){2-3} \cmidrule(lr){4-5} \cmidrule(lr){6-7} \cmidrule(lr){8-9} 
\cmidrule(lr){10-11} \cmidrule(lr){12-13} \cmidrule(lr){14-15} \cmidrule(lr){16-17}
& MAE & MSE & MAE & MSE & MAE & MSE & MAE & MSE 
& MAE & MSE & MAE & MSE & MAE & MSE & MAE & MSE & \\
\midrule
DLinear       & 0.375 & 0.278 & 0.388 & 0.295 & 0.396 & 0.329 & 0.416 & 0.379 & 0.268 & 0.229 & 0.292 & 0.203 & 0.260 & 0.299 & 0.357 & 0.248 & 0 \\
MegaCRN       & 0.263 & 0.155 & 0.243 & 0.129 & 0.239 & 0.146 & 0.254 & 0.206 & 0.244 & 0.222 & 0.286 & 0.186 & 0.318 & 0.225 & 0.263 & 0.157 & 0 \\
Memformer     & 0.259 & 0.146 & 0.269 & 0.154 & 0.241 & 0.129 & 0.273 & 0.176 & 0.242 & 0.223 & 0.270 & 0.184 & 0.230 & \underline{0.176} & 0.266 & \underline{0.155} & 0 \\
iTransformer  & 0.222 & 0.113 & 0.221 & 0.111 & 0.204 & 0.101 & 0.226 & 0.150 & 0.237 & 0.222 & 0.255 & 0.166 & 0.225 & 0.192 & 0.284 & 0.171 & 0 \\
TimeLLM       & 0.275 & 0.168 & 0.290 & 0.188 & 0.289 & 0.202 & 0.306 & 0.247 & 0.241 & 0.225 & 0.280 & 0.196 & 0.238 & 0.190 & 0.286 & 0.174 & 0 \\
ModernTCN     & 0.233 & 0.128 & 0.229 & 0.116 & 0.218 & 0.118 & 0.267 & 0.179 & 0.240 & 0.223 & 0.286 & 0.197 & 0.269 & 0.198 & 0.275 & 0.166 & 0 \\
STONE         & 0.277 & 0.163 & 0.269 & 0.151 & 0.270 & 0.166 & 0.281 & 0.209 & 0.239 & 0.213 & 0.297 & 0.201 & 0.257 & 0.238 & 0.332 & 0.228 & 0 \\
TimeKAN       & 0.303 & 0.198 & 0.314 & 0.217 & 0.317 & 0.239 & 0.323 & 0.261 & \underline{0.232} & \textbf{0.209} & 0.277 & 0.185 & 0.265 & 0.216 & 0.283 & 0.173 & \underline{1} \\
TimePro       & 0.245 & 0.137 & 0.228 & 0.118 & 0.250 & 0.156 & 0.262 & 0.196 & 0.238 & 0.217 & \underline{0.248} & \underline{0.153} & 0.231 & 0.191 & \underline{0.258} & \textbf{0.150} & \underline{1} \\
FilterTS      & 0.249 & 0.144 & 0.239 & 0.129 & 0.223 & 0.142 & 0.252 & 0.185 & \underline{0.232} & \underline{0.211} & 0.256 & 0.164 & 0.239 & 0.191 & 0.265 & 0.160 & 0 \\
TimeMixer++   & 0.225 & 0.209 & 0.248 & 0.136 & 0.241 & 0.137 & 0.270 & 0.185 & 0.238 & 0.214 & 0.261 & 0.167 & \underline{0.211} & 0.207 & 0.268 & 0.157 & 0 \\
DUET          & \underline{0.214} & \underline{0.109} & \underline{0.218} & \underline{0.110} & \underline{0.192} & \underline{0.090} & \underline{0.214} & \underline{0.120} & 0.244 & 0.216 & 0.252 & \underline{0.153} & 0.222 & \textbf{0.165} & 0.272 & 0.159 & \underline{1} \\
\midrule
\multirow{2}{*}{\textbf{VLBM}} 
& \textbf{0.191} & \textbf{0.092} & \textbf{0.188} & \textbf{0.087} & \textbf{0.167} & \textbf{0.076} & \textbf{0.186} & \textbf{0.104} & \textbf{0.225} & \textbf{0.209} & \textbf{0.240} & \textbf{0.152} & \textbf{0.194} & 0.181 & \textbf{0.255} & \textbf{0.150} & \multirow{2}{*}{\textbf{15}} \\
& \color{gray}\tiny($\pm$0.0003) & \color{gray}\tiny($\pm$0.0003) 
& \color{gray}\tiny($\pm$0.0002) & \color{gray}\tiny($\pm$0.0003) 
& \color{gray}\tiny($\pm$0.0002) & \color{gray}\tiny($\pm$0.0004) 
& \color{gray}\tiny($\pm$0.0004) & \color{gray}\tiny($\pm$0.0013) 
& \color{gray}\tiny($\pm$0.0013) & \color{gray}\tiny($\pm$0.0015) 
& \color{gray}\tiny($\pm$0.0003) & \color{gray}\tiny($\pm$0.0009) 
& \color{gray}\tiny($\pm$0.0004) & \color{gray}\tiny($\pm$0.0007) 
& \color{gray}\tiny($\pm$0.0003) & \color{gray}\tiny($\pm$0.0002) & \\
\bottomrule
\end{tabular}
\end{table*}

\section{Experiment}
Our experiments examine whether VLBM improves forecasting reliability under
mixed ID/OOD conditions and whether its latent structure behaves as intended.
We first evaluate accuracy on standard ID benchmarks and anomalous OOD
intervals. We then inspect the learned basis, the residual representation,
and posterior prior transfer, followed by component and capacity analyses.
We investigate four questions.\\
\textbf{Q1 (Forecasting Reliability).} Whether VLBM improves forecasting accuracy under both standard ID conditions and rare, high impact OOD intervals.\\
\textbf{Q2 (Latent Space Separation).} Whether the learned low rank latent space captures stable ID dynamics while isolating OOD deviations.\\
\textbf{Q3 (Posterior Prior Transfer).} Whether aligning a \emph{future aware posterior} with a \emph{future blind prior} distills future informed latent dynamics into test time inference and improves robustness under distribution shift.\\
\textbf{Q4 (Component and Capacity Analysis).} How VLBM's major components and capacity related hyperparameters affect OOD robustness and forecasting stability.

\subsection{Experimental Settings}
\subsubsection{Datasets} We evaluate VLBM on 12 public multivariate datasets spanning standard forecasting benchmarks and anomaly-labeled datasets for out-of-distribution (OOD) evaluation.
For in-distribution (ID) evaluation \cite{TimesNet,itransformer,MSGNet}, we use (1) \textit{Weather}: Monitoring 21 meteorological parameters from the Max Planck Institute in Germany. (2) \textit{ECL}: Updated electricity consumption data from 321 users. (3) \textit{Solar}: Irradiance data from photovoltaic solar power stations. (4) \textit{Flight}: Flight data from seven major European airports. (5) \textit{PEMS}: Traffic flow data collected at high speed by the Caltrans Performance Measurement System.
For OOD robustness evaluation, we include anomaly-labeled datasets and construct two incident-aligned traffic datasets in CHP-LCS: (1) \textit{MSL} \cite{hundman2018detecting}: Abnormal Events and Related Data of NASA Mars Exploration Mission ``Curiosity'' Rover. (2) \textit{PSM} \cite{abdulaal2021practical}: eBay's IT system signal dataset, which detects server node anomalies based on generated signals. (3) \textit{CHP-LCS} (Speed and Flow): We follow the PeMS anomaly labeling setting in \cite{yichang}, and align PeMS Bay traffic measurements with California Highway Patrol incident reports and Lane Closure System reports. The resulting datasets contain 365 sensors with 5 min flow and speed series, where flow is aggregated across lanes and speed is averaged across lanes. The graph is built from sensor adjacency and inverse distance weights with row normalization.

\subsubsection{Baselines} 12 state of the art baselines are compared:

\begin{itemize}
\item \textit{Transformer based:} DUET \cite{DUET} employs temporal clustering to handle heterogeneous distribution shifts and frequency domain channel soft clustering to capture channel interactions while filtering noise. Memformer \cite{Memformer} uses memory enhanced attention and global memory tokens in a dual branch architecture to capture historical trends and local fluctuations. iTransformer \cite{itransformer} applies dimension wise tokenization to embed variables independently, utilizing attention for multivariate correlations.

\item \textit{MLP based:} TimeMixer++ \cite{wang2025timemixer++} transforms multi scale time series into multi resolution images, leveraging hierarchical mixing across time scales and frequency resolutions for patterns adaptation. FilterTS \cite{filterts} employs adaptive frequency decomposition and filtering to dynamically enhance key features and suppress noise. DLinear \cite{dlinear} decomposes sequences into trend and seasonal components, applying simple linear layers to reduce complexity.

\item \textit{Memory based:} STONE \cite{Stone} tackles spatio temporal shifts by combining Fréchet embeddings with a graph intervention to learn invariant node representations. MegaCRN \cite{megacrn} introduces Spatio Temporal Meta Graph Learning to effectively disentangle heterogeneous spatio temporal patterns.

\item \textit{LLM based:} Time-LLM \cite{Time-LLM} converts time series into a language space via patch embedding and uses prompt prefixes to activate the LLM reasoning capabilities.

\item \textit{TCN based:} ModernTCN \cite{ModernTCN} upgrades TCNs with large kernel convolutions to expand receptive fields and separation convolutions to decouple variable correlation.

\item \textit{Mamba based:} TimePro \cite{timepro} uses a time tune strategy and reconstructed hyper states to efficiently model variable relationships and temporal dynamics.

\item \textit{KAN based:} TimeKAN \cite{TIMEKAN} uses discrete Fourier transforms to capture dynamic patterns and fluctuations.

\end{itemize}

Additionally, some baselines inherently support OOD shifts: Memformer and STONE extract invariant causal dependencies via memory-augmented graph learning, while ModernTCN uses effective receptive fields for anomaly detection.

\textbf{Setup.}
All experiments are implemented in PyTorch and conducted on NVIDIA GeForce RTX 4090 GPU.
We report MAE and MSE as evaluation metrics.
Dataset splits follow standard protocols in prior work. PEMS datasets use a 6:2:2 training, validation, and test ratio, and other datasets use a 7:1:2 ratio.
For OOD settings, evaluation is performed exclusively on anomalous test intervals.
This evaluation measures robustness to rare shifts whose precursors or initial deviations are present in the historical window \(X\). VLBM receives no future event labels during training or inference. Its goal is to preserve and extrapolate weak historical deviation signals that may be smoothed out by models dominated by frequent ID patterns.
All models adopt a unified look back window of 96 time steps for fair comparison.
Complete horizon-wise results, additional visualizations, efficiency comparisons,
and implementation details are provided in the supplementary material.

\begin{figure}
\centering
\includegraphics[width=\linewidth]{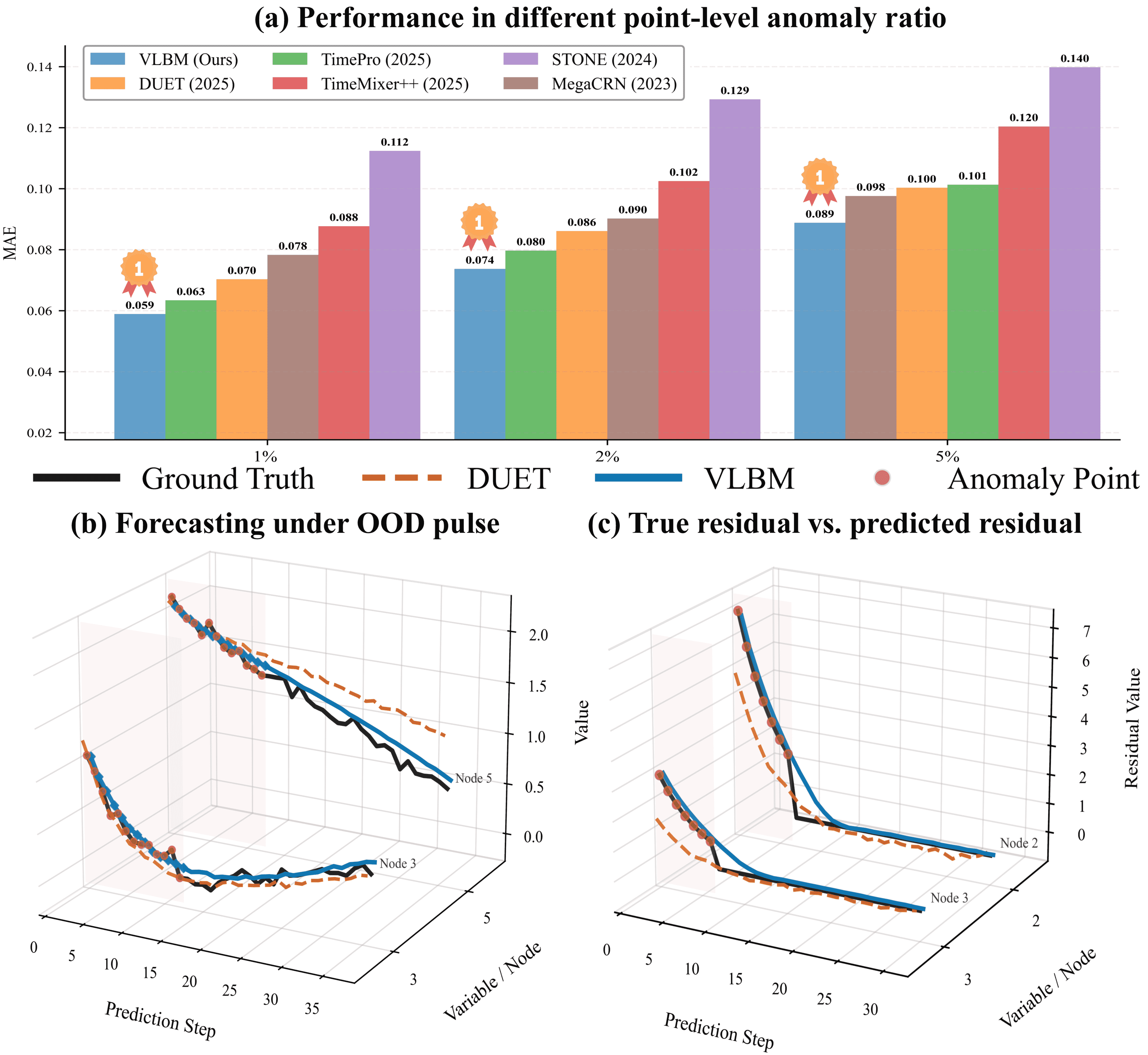}
\caption{
Controlled simulation on the Synthetic Graph Pulse benchmark.
(a) Forecasting MAE under 1\%, 2\%, and 5\% point level anomaly ratios.
(b) Forecasting comparison under an OOD pulse interval.
(c) Ground truth OOD residual versus the residual response produced by VLBM.
}
\label{fig:simulation}
\end{figure}

\begin{figure*}[!t]
\centering
\includegraphics[width=\linewidth]{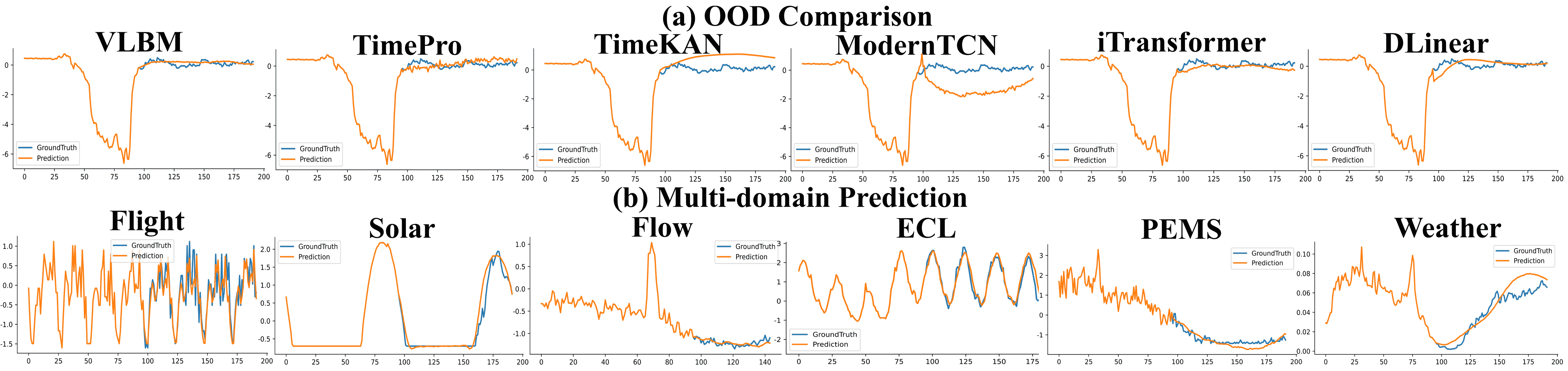}
\caption{
Forecasting comparison under ID and OOD conditions.
(a) Forecasting results on the CHP-LCS-Speed dataset, evaluated on anomalous intervals corresponding to real world incidents.
(b) Forecasting results on ID datasets with diverse temporal dynamics.
}
\label{fig:vis}
\end{figure*}

\subsection{Performance under ID and OOD Conditions (RQ1)}

\textbf{Comparison Analysis.}
Tables~\ref{tab:ood} and~\ref{tab:id} report MAE and MSE averaged across multiple forecasting horizons on OOD and ID datasets, respectively.
Across 12 datasets and two metrics (24 averaged evaluations), VLBM outperforms the strongest baseline in 22 cases.

Under OOD settings, evaluation is performed only on time intervals annotated as anomalous to assess robustness under distribution shift.
No anomaly labels are used during training. They are only used to select OOD intervals for evaluation.
As shown in Table~\ref{tab:ood}, VLBM achieves the best performance on all OOD datasets, with consistent gains over strong baselines.
This suggests that VLBM goes beyond average case optimization and better handles rare but high impact distribution shifts. Under standard ID settings, Table~\ref{tab:id} shows that VLBM achieves state of the art forecasting accuracy on
most benchmarks, with consistently strong performance across all settings.
These results indicate that VLBM enhances OOD robustness and preserves
strong in distribution forecasting performance.

\textbf{Controlled Simulation on Synthetic Graph Pulse.}
To verify whether VLBM separates stable dynamics from rare OOD deviations, we construct a Synthetic Graph Pulse (SGP) benchmark. SGP generates 40 node time series with daily and weekly periodic patterns over a fixed graph, and injects positive or negative pulse residuals as controlled anomalies. Each pulse starts from a source node, exponentially decays back to the normal regime, and propagates with weakened amplitudes to one hop and two hop neighbors. The synthetic setting provides ground truth residuals that are unavailable in real datasets, so it helps check whether the residual path responds to injected OOD pulses.

We evaluate three point level anomaly ratios: 1\%, 2\%, and 5\%, corresponding to 119, 225, and 594 injected events. As shown in Fig.~\ref{fig:simulation}(a), VLBM performs best across all anomaly ratios, indicating robust performance as OOD pulses become more frequent. Fig.~\ref{fig:simulation}(b) shows that VLBM stays closer to the ground truth than DUET in both anomalous and regular regions, better tracking pulse deviations and recovery. Using the available ground truth residuals in Fig.~\ref{fig:simulation}(c), we further observe that VLBM closely matches the true residual and gradually returns to normal. DUET mainly follows regular patterns and shows weaker responses to anomaly residuals. This confirms that VLBM’s residual path is effectively activated around OOD deviations.

\textbf{Qualitative Analysis.}
Fig.~\ref{fig:vis} provides qualitative comparisons of forecasting
performance under both OOD and ID conditions.
Fig.~\ref{fig:vis}(a) shows predictions on the CHP-LCS-Speed dataset, where real world incidents induce abrupt distribution shifts.
Under these anomalous conditions, VLBM closely tracks the ground truth, demonstrating strong robustness to OOD perturbations.
Baseline models follow regular patterns but show weaker adaptation to OOD dynamics.
This limitation is consistent with the rare OOD suppression effect in mixture risk training, where OOD signals have limited influence compared with frequent in distribution patterns. Fig.~\ref{fig:vis}(b) presents forecasting results across multiple
domains with diverse temporal characteristics, including stable solar
generation and highly fluctuating flight traffic.
Across all scenarios, VLBM produces predictions closely aligned with the
ground truth, demonstrating strong adaptability under heterogeneous dynamics.

\subsection{Analysis of the Latent Space (RQ2)}
\begin{figure}[!t]
\centering
\includegraphics[width=0.9\linewidth]{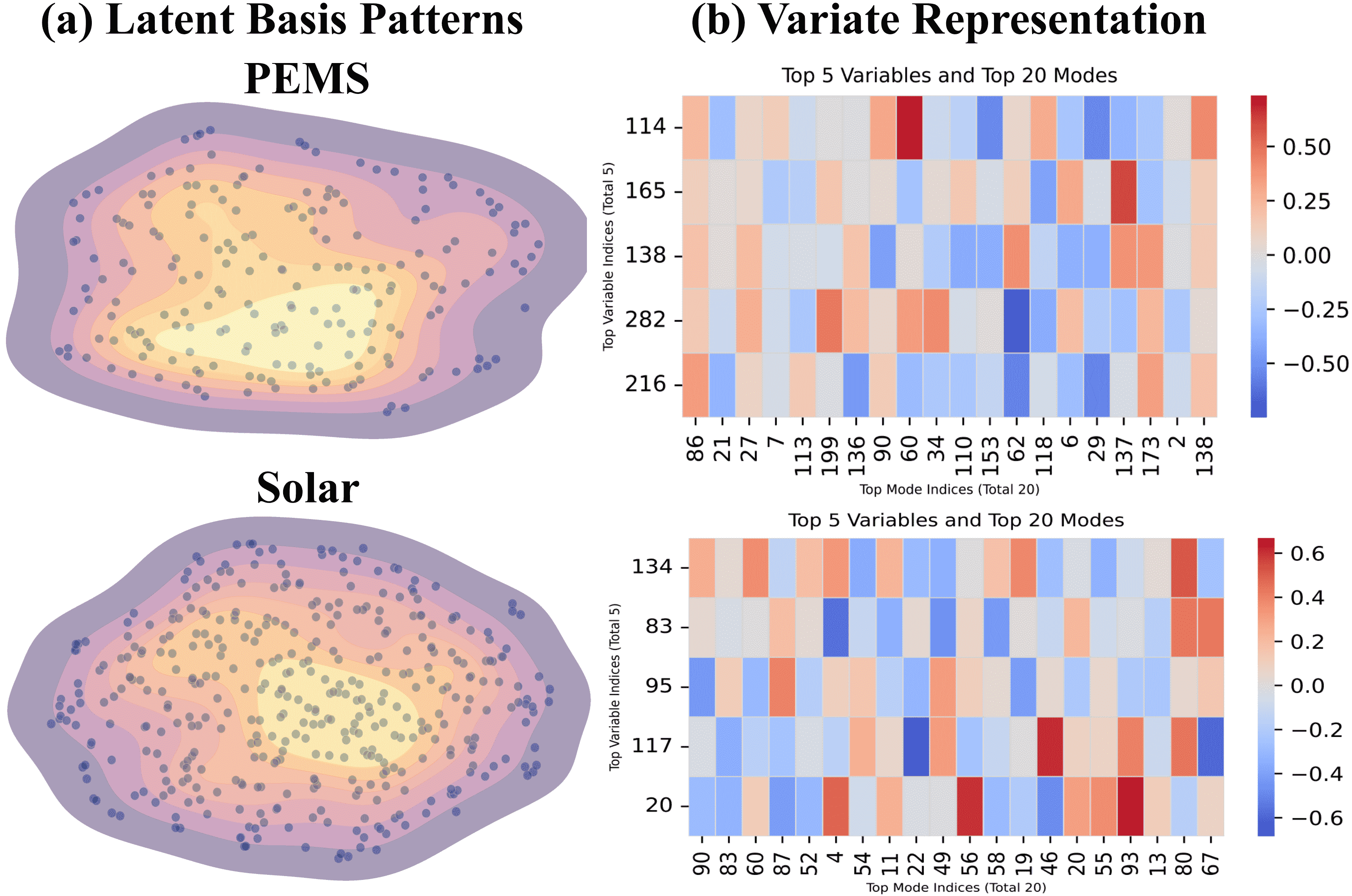}
\caption{
Visualization of learned latent basis.
(a) UMAP visualization of latent basis vectors.
(b) Activation patterns of representative variables over the most strongly activated latent basis.
}
\label{fig:lbv}
\end{figure}

\textbf{Latent Basis Patterns Visualizations.}
Latent basis vectors are designed to capture shared structure, such that diverse patterns in the data can be represented as weighted combinations of the latent basis.
To analyze their structure, we visualize
latent basis patterns on the PEMS and Solar datasets using UMAP. Fig.~\ref{fig:lbv}(a) shows that the learned latent basis are geometrically separated in the projected space.
Fig.~\ref{fig:lbv}(b) visualizes the activation patterns of representative variables over their most strongly activated basis vectors. Different variables activate distinct basis combinations, suggesting that the bases provide temporal structural primitives that variables selectively compose beyond their geometric separation in a low dimensional visualization.
Together, these results provide empirical evidence that the shared latent basis
captures in distribution structure and provides a structural foundation for decoupling stable dynamics from deviation components.

\begin{figure}
\centering
\includegraphics[width=0.9\linewidth]{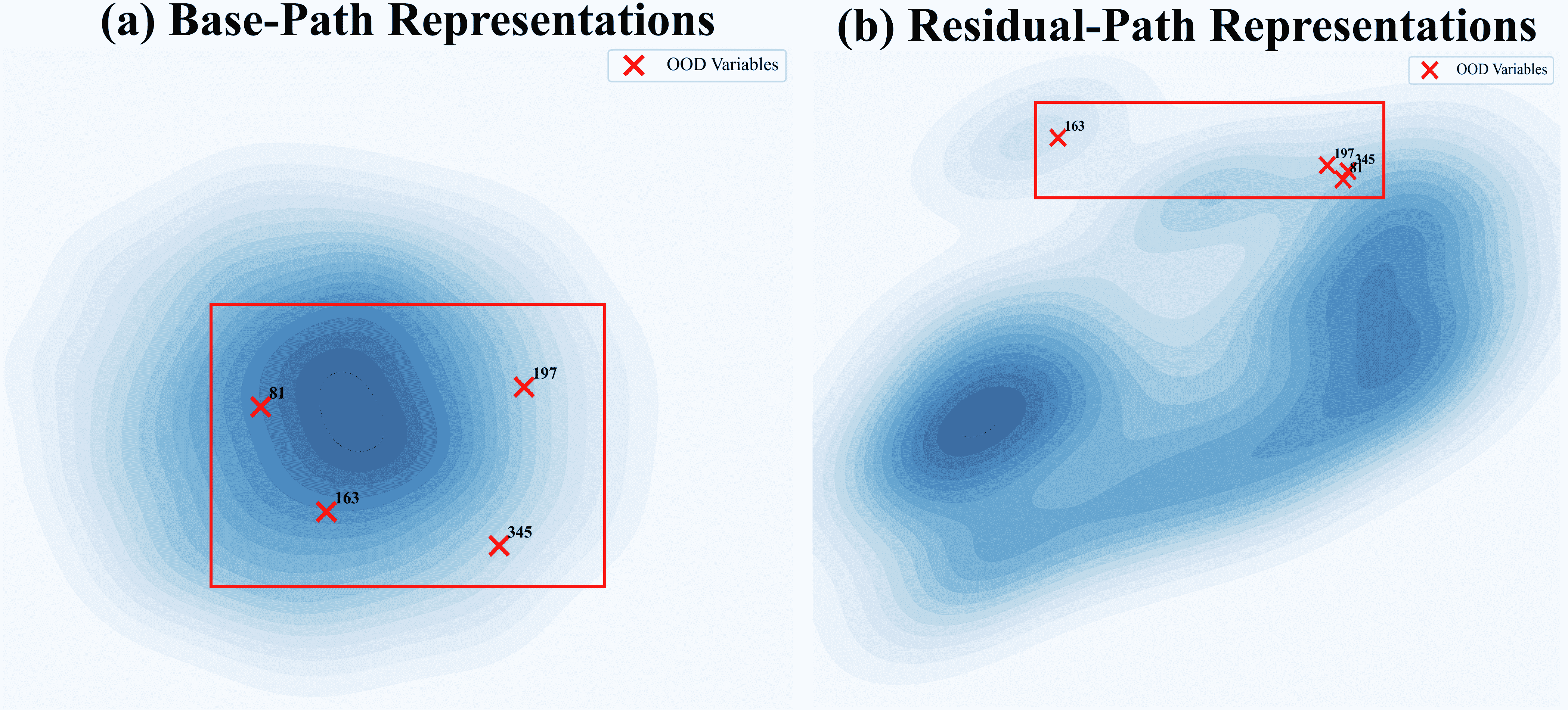}
\caption{
Visualization of pathway representations.
(a) Base path representation for stable dynamics.
(b) Residual path representation for deviation components.
}
\label{fig:lsv}
\end{figure}
\textbf{Latent Space Decomposition.}
Using randomly sampled one-hour time slices from the CHP-LCS-Flow dataset,
we visualize the representations used by the Base and Residual paths.
The Base path representation is constructed from the prior-induced latent
state \(\mathbf Z^{(p)}\) and the stable projected component
\(X_{\parallel}^{(B)}\), while the Residual path representation is constructed
from the orthogonal component \(X_{\perp}^{(B)}\) and the historical
representation \(X\). Both representations are projected into two dimensions
using t-SNE for qualitative visualization.

As shown in Fig.~\ref{fig:lsv}(a), anomalous samples largely overlap with
the overall ID-dominated distribution in the Base path space, suggesting
that the Base path preserves shared stable dynamics rather than amplifying
anomaly-specific variations. In contrast, Fig.~\ref{fig:lsv}(b) shows that
OOD samples are more dispersed and geometrically irregular in the Residual
path space, while ID samples remain compact. This indicates that OOD-induced
variations are mainly expressed through the residual channel rather than
absorbed into the stable Base representation.

\begin{figure}
\setlength{\abovecaptionskip}{1pt}
\centering
\includegraphics[width=\linewidth]{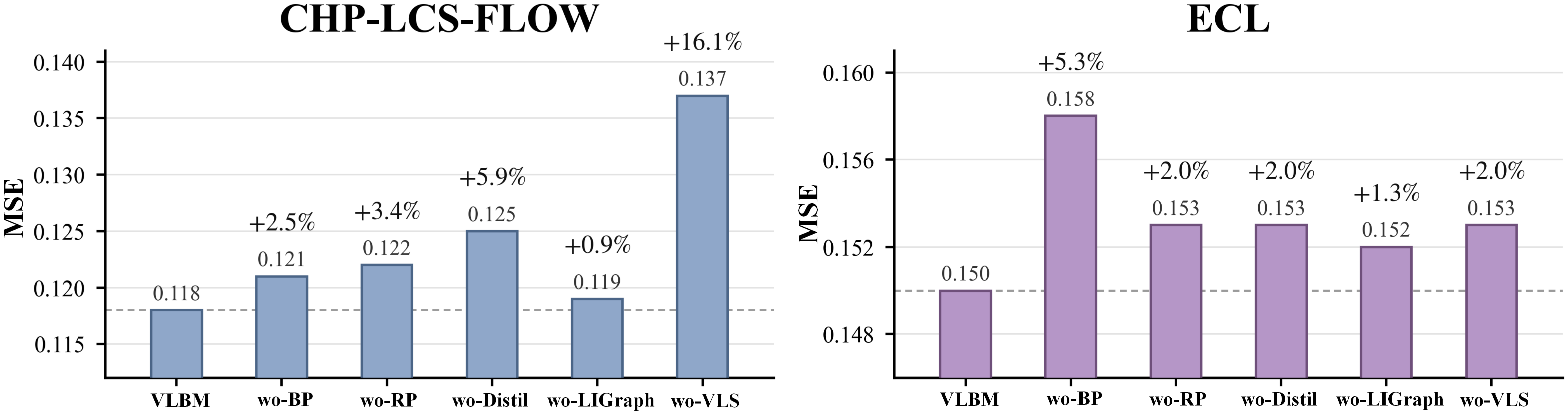}
\caption{Ablation study on CHP-LCS-Flow (OOD) and ECL (ID) in averaged forecast horizons. All removals are based on the optimal training settings.}
\label{fig:ab}
\end{figure}

\subsection{Ablation Studies and Analysis (RQ3, RQ4)}

We conduct ablation studies on VLBM's core components to assess their
individual and joint contributions under varying distributional
conditions.
Specifically, \textit{wo-BP}, \textit{wo-RP}, \textit{wo-Distil},
\textit{wo-LIGraph}, and \textit{wo-VLS} remove the Base Path, Residual
Path, posterior prior transfer, latent induced graph, and
latent space module, respectively.
We evaluate these variants on two representative datasets spanning
both ID and OOD scenarios, with results summarized in Fig.~\ref{fig:ab}.

\textbf{Effect of Variational Latent Modeling.}
Under the CHP-LCS-Flow OOD scenario, removing the variational latent
space module (\textit{wo-VLS}) results in the most pronounced performance
degradation. The \textit{wo-VLS} variant removes the whole latent space module,
including latent states, basis induced projection, and orthogonal residual
decomposition. The Base and Residual paths are then built directly from the
original historical representation \(X\). Its degradation under CHP-LCS-Flow
supports the role of latent structured modeling in organizing stable dynamics
and deviation specific information.
In addition, removing posterior prior transfer
(\textit{wo-Distil}) also leads to a notable performance drop,
demonstrating the importance of distilling future aware latent
activations into a future blind prior to enable stable test time latent
inference.

\textbf{Effect of Base--Residual Decomposition.}
On the regular ECL dataset, removing the Base Path (\textit{wo-BP})
results in the most significant performance degradation.
This indicates that the Base Path plays a dominant role in capturing
stable and reusable in distribution dynamics under standard forecasting
conditions.
The remaining components yield smaller but consistent performance
gains, indicating their complementary contributions.

\textbf{Effect of Module Interdependence.} Removing any major component leads to a substantial
increase in forecasting error.
This indicates strong interdependence among VLBM’s components under both OOD or ID conditions.
The latent induced graph provides an auxiliary propagation mechanism for
the residual path, while the main robustness gains come from the
interaction between variational latent modeling and Base--Residual
decomposition.

\begin{table}[!t]
\setlength{\abovecaptionskip}{1pt}
  \centering
  \caption{Optimal analysis of hyperparameter $M$ (MSE)}
  \label{tab:sensitivity_m}
  \small 
\resizebox{\linewidth}{!}{%
  \begin{tabular}{lccccc}
    \toprule
    Dataset &$M=100$ & $M=150$ & $M=200$ & $M=300$ & $M=500$ \\
    \midrule
    PEMS03 & 0.1316 & 0.1329 & 0.1325 & \textbf{0.1307} & 0.1324 \\
    MSL    & 6.3581 & 6.3837 & 6.3803 & \textbf{6.3561} & 6.3689 \\
    \bottomrule
  \end{tabular}
}

\end{table}
\subsection{Hyperparameter Analysis}

We analyze the sensitivity of VLBM to the number of latent basis $M$ and the neighborhood size $K$ in the latent induced graph.
The parameter $M$ controls the capacity of the low rank latent subspace,
i.e., the number of reusable in distribution patterns.
On both PEMS03 and MSL, the best performance is achieved around
$M=300$, indicating that while the underlying dynamics are rich, they
can be effectively captured by a finite number of latent modes.
Smaller values of $M$ limit expressiveness, and overly large $M$
weakens the low rank constraint and may reduce the separation between
stable structure and deviation specific information. This trend is
consistent with the capacity separation tradeoff discussed in
Section~\ref{subsec:low-rank-residual}.

\section{Conclusion}
This work studies multivariate time series forecasting under mixed
ID and OOD conditions, where rare but high impact events are often
overshadowed by representation learning dominated by frequent ID patterns.
We propose \textbf{VLBM}, which moves forecasting into a latent space
by learning a shared low-rank latent subspace to capture transferable
in-distribution dynamics while separating OOD-induced deviations.
A variational posterior prior transfer scheme aligns a \emph{future-aware posterior}
with a \emph{future-blind prior} for stable test-time inference,
and a Base--Residual generator further decouples stable dynamics
from deviation propagation.
Our experiments suggest that robust forecasting is supported by
disentangling stable latent dynamics from rare, distribution specific
deviations and by avoiding a single observation space representation for
all observed variation.
Across 12 real world benchmarks, including newly constructed OOD datasets,
VLBM achieves state of the art OOD robustness and in distribution accuracy.
Results on a synthetic simulation dataset further show that VLBM can track
controlled OOD pulse deviations and recovery.
These results support latent structured forecasting as a principled route
for improving reliability under mixed ID and OOD conditions.
This study also has limitations.
The stable subspace assumption may be imperfect in systems where regular
dynamics and OOD deviations are strongly coupled, and VLBM does not explicitly
use external event or weather inputs.
Future work may combine latent structured forecasting with adaptive basis
regularization and uncertainty diagnostics.

\section*{Acknowledgments}
\noindent This study was supported by the National Natural Science Foundation of China (Grant No. 52071312).

\section*{Data and Code Availability}
\noindent The public datasets used in this study are available from their original sources cited in this article. The source code and newly constructed real-world OOD datasets for reproducing the experiments are available at \href{https://github.com/leijieruilq/VLBM_OOD_forecast}{https://github.com/leijieruilq/VLBM\_OOD\_forecast}.

\bibliographystyle{IEEEtran}
\bibliography{ref}

\end{document}